\documentclass[12pt]{article}
\usepackage[english]{babel}
\usepackage[utf8]{inputenc}
\usepackage[T1]{fontenc}
\usepackage{lmodern}
\usepackage[table]{xcolor}
\usepackage{johd}
\usepackage{fancyhdr}
\fancypagestyle{preprint}{%
  \fancyhf{}%
  \fancyhead[L]{\small Preprint. Under review.\vspace{2pt}\hrule}%
  \fancyfoot[C]{\thepage}%
}
\usepackage{microtype}
\usepackage{booktabs}
\usepackage{graphicx}
\usepackage{amsfonts}
\usepackage{amsmath}
\usepackage{nicefrac}
\usepackage{tcolorbox}
\tcbuselibrary{breakable}

\definecolor{darkblue}{rgb}{0, 0, 0.5}
\hypersetup{colorlinks=true, citecolor=darkblue, linkcolor=darkblue, urlcolor=darkblue, hypertexnames=false}

\newcommand{\proj}{AutoTailor}

\newtcolorbox{promptbox}{
  colback=gray!5,
  colframe=gray!50,
  boxrule=0.5pt,
  arc=2pt,
  left=6pt,
  right=6pt,
  top=6pt,
  bottom=6pt,
  breakable
}

\title{\proj{}: Automatic, User-Aligned Capability Selection and Adaptation for Web Agents}
\author{%
  Xinyun Cao$^1$, Adriana Szekeres$^2$, Fazle Elahi Faisal$^2$ \\
  \small $^1$University of Michigan, Ann Arbor \\
  \small $^2$Microsoft Research, Redmond, WA \\
  \small \texttt{xinyunc@umich.edu, \{aszekeres,fafaisal\}@microsoft.com}
}
\date{}

\begin{document}

\maketitle
\thispagestyle{preprint}

\begin{abstract}
Web agents can utilize reusable tools to reduce the cost and latency of low-level browser interaction, but automatically discovered tool collections can be large, redundant, and poorly aligned with user demand. We present \proj{}, a meta-agentic framework for constructing and maintaining a compact set of trajectory-derived Model Context Protocol (MCP) APIs. Offline, \proj{} converts web trajectories into parameterized browser-automation programs, applies a Quality Filter to remove APIs with unsuitable granularity and redundant functionality, and applies a Usage Likelihood Filter to prioritize broadly useful capabilities while preserving semantic coverage. Online, Dynamic Reselection monitors task outcomes and API usage, identifies recurring coverage gaps, adds relevant candidates, and prunes persistently unused capabilities. We evaluate \proj{} on 106 WebArena Postmill tasks. Offline filtering reduces the initial 1,283 unrefined APIs to 87, and Dynamic Reselection produces a 33-API set. With reasoning and acting (ReAct) fallback, this set achieves 90.6\% correctness, compared with 87.5\% for ReAct alone, while reducing average total request-token cost by 57.8\% and latency by 29.4\%. Without ReAct, it achieves 60.1\% correctness, marginally matching the performance of unrefined set, while reducing request-token usage by 94.9\%. Together, these results show that static filtering produces a compact inventory of APIs expected to support core, high-likelihood tasks, while dynamic reselection further tailors that inventory to observed user needs. This combination improves accuracy and latency while sharply reducing token usage and end-to-end cost, demonstrating the value of user-aligned capability management for efficient web agents.
\end{abstract}

\begin{figure}[!t]
  \centering
  \includegraphics[width=\textwidth]{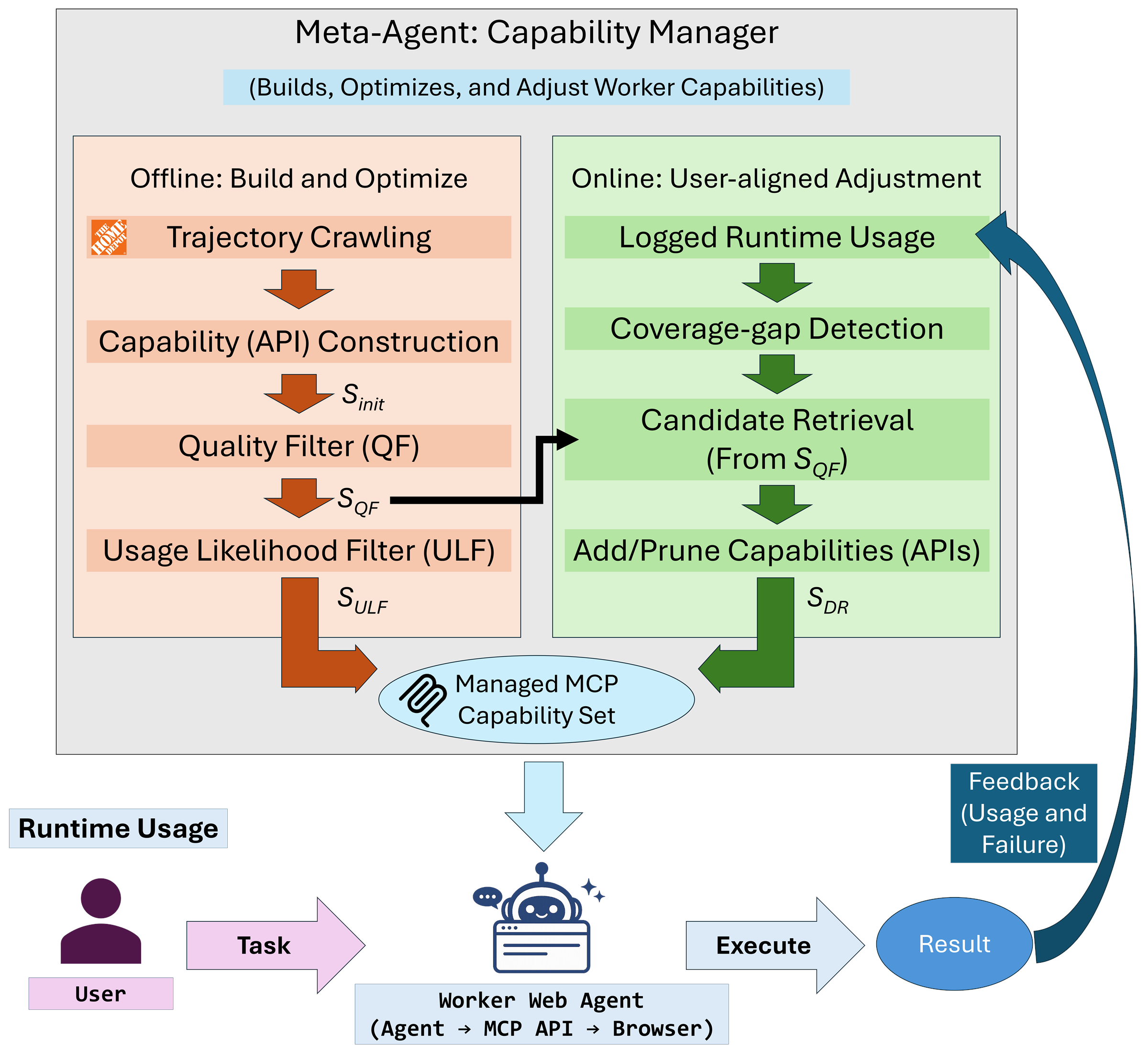}
  \caption{\textbf{Overview of \proj{} system architecture:} trajectory-derived web actions are converted into reusable MCP APIs, filtered and deduplicated into a compact capability bank, and then dynamically reselected to match user demand while maintaining functional coverage for downstream runtime web agents.}
  \label{fig:system_diagram}
\end{figure}

\section{Introduction}
\label{sec:intro}
A web agent is an autonomous AI system that perceives web content, interprets user goals expressed through natural-language instructions, and interacts with web interfaces through actions such as clicking, typing, and scrolling to complete tasks on a user's behalf \cite{ning2025surveywebagentsnextgenerationai}. Such systems have the potential to support a broad range of applications, from improving the efficiency of interactions with complex websites \cite{ning2025surveywebagentsnextgenerationai} to increasing web accessibility through non-visual interaction modalities \cite{colazzo2026zoominginagenticweb}.

Early web agents commonly followed the ReAct interaction paradigm, repeatedly observing a screenshot or Document Object Model (DOM) tree, reasoning about the next action, executing that action, and reassessing the resulting webpage state until task completion. This iterative process, however, can be error-prone, slow, and computationally expensive \cite{sodhi2024stepstackedllmpolicies,fourney2024magentic,yang2024agentoccamsimplestrongbaseline,zhang2025symbioticcooperationwebagents,openai2024cua}. Recent systems increasingly employ reusable skills and tools to improve the speed, accuracy, and cost-effectiveness of web-agent deployment \cite{zheng2025skillweaver,prabhu2025walt,wang2026webxskill}. Nevertheless, determining how to construct a web agent's ``capability space'' so that it remains efficient while aligning with user needs is an open problem \cite{lou2026toolillusion}. Manually constructing skills and tools is costly \cite{schick2023toolformer}, whereas automated discovery systems may provide insufficient coverage \cite{pahuja2025explorer,zheng2025skillweaver, prabhu2025walt}. Prior work has introduced automated agentic web crawling to improve coverage at lower cost by discovering large collections of candidate behaviors \cite{faisal2026autosurfer}. However, these behaviors may be redundant, overlapping, or poorly aligned with user needs.

These limitations motivate a reliable framework for constructing a compact yet comprehensive web-agent capability space represented as MCP APIs. We propose \proj{}, a meta-agentic framework with two main capability-management modules: an \textbf{Offline Selection module} and an \textbf{Online Dynamic Reselection module} (see Figure \ref{fig:system_diagram}). The offline module constructs the initial capability set through three stages: \emph{Trajectory Crawling}, which collects diverse web interaction trajectories; \emph{Capability (API) Construction}, which converts trajectories into reusable parameterized programs; and \emph{Capability Selection}, which applies a Quality Filter (QF) followed by a Usage Likelihood Filter (ULF) to produce the initial managed MCP capability set. At runtime, the online module adapts the active capability set through four stages: \emph{Logged Runtime Usage}, which records user intents, tool usage, and task outcomes; \emph{Coverage-Gap Detection}, which identifies recurring tasks that are not adequately supported by the current APIs; \emph{Candidate Retrieval} from a candidate API pool, which selects inactive APIs that can address the detected gaps; and \emph{Capability Addition and Pruning}, which incorporates useful candidates while removing persistently unused APIs. This design combines broad offline capability discovery with usage-driven online adaptation, allowing the deployed tool set to remain compact while tracking the capabilities required by the observed task distribution.

We evaluate \proj{} on 106 tasks from the Postmill environment of WebArena \cite{zhou2024webarenarealisticwebenvironment}, comparing the initial unrefined ($S_{\mathrm{init}}$), statically selected (after QF forms $S_{\mathrm{QF}}$, after ULF forms $S_{\mathrm{ULF}}$), and dynamically reselected ($S_{\mathrm{DR}}$) API sets with and without a ReAct fallback method. We measure task success, latency, API usage, and model-invocation and token costs. Our evaluation provides insight into the tradeoffs among capability-set size, efficiency, and functional coverage.
QF and ULF reduce the capability pool from $|S_{\mathrm{init}}|=1{,}283$ APIs to $|S_{\mathrm{ULF}}|=87$, while DR further adapts it to the compact $|S_{\mathrm{DR}}|=33$ set. Our results show that, with the ReAct Fallback enabled, compared to using ReAct alone, $S_{\mathrm{DR}}$ improves correctness from 87.5\% to 90.6\% while reducing average total request-token cost by 57.8\% and latency by 29.4\%. Without ReAct, $S_{\mathrm{DR}}$ retains 60.1\% correctness, marginally matching the performance of $S_{\mathrm{init}}$, while reducing request-token usage by 94.9\%. These results show that \proj{} can substantially compress the deployed capability set while preserving, and in the ReAct-enabled setting improving, task performance.

Our contributions are as follows:
\begin{itemize}
  \item We present \proj{}, a meta-agentic framework for constructing and maintaining a compact portfolio of executable web capabilities that balances functional coverage, deployment efficiency, and alignment with observed user needs.

  \item We show that offline capability selection can substantially reduce the size and tooling overhead of the candidate API pool, while revealing a tradeoff between compactness, direct functional coverage, and reliance on ReAct fallback.

  \item We show that dynamic reselection mitigates this tradeoff by adapting the deployed capability set to observed runtime demand, recovering coverage with a smaller capability set and improving efficiency and task performance both with and without ReAct fallback.
\end{itemize}

\section{Related Work}
\subsection{Web trajectory generation}

Large-scale, environment-grounded trajectories provide the raw behaviors from which web-agent capabilities can be derived. Prior work has expanded this method along complementary dimensions. OS-Genesis reverses conventional task-first collection by exploring a GUI first, synthesizing tasks retrospectively, and applying a trajectory reward model for quality control~\cite{sun2025osgenesis}. Explorer combines broad web exploration with intent refinement to generate successful multimodal trajectories at scale~\cite{pahuja2025explorer}. To improve the reliability of the resulting data, SynthAgent jointly refines synthetic tasks and demonstrations to reduce infeasible instructions and noisy or misaligned actions~\cite{wang2026synthagent}, while AutoSurfer couples systematic breadth-first exploration with trajectory-grounded task synthesis and refinement to increase website coverage~\cite{faisal2026autosurfer}. These methods focus on obtaining broader, more diverse, or more reliable trajectory corpora. Our work takes such a corpus as input and addresses the next question: which behaviors represented in thousands of trajectories should become deployed tools? Specifically, \proj{} identifies a compact set of high-value capabilities while excluding candidates that are redundant, overly specific, poorly formed, or unlikely to serve user requests.

\subsection{Tool and skill synthesis}

Trajectory data become directly reusable when low-level interactions are abstracted into parameterized tools or skills. Existing systems explore several paths to this abstraction. Agent Skill Induction (ASI) composes primitive actions into higher-level routines and verifies the induced programs online~\cite{wang2025asi}. SkillWeaver discovers and practices website interactions, then distills them into a growing library of reusable APIs~\cite{zheng2025skillweaver}. WALT instead recovers functionality already present in a website and exposes it as invocable tools~\cite{prabhu2025walt}, whereas WebXSkill mines reusable subsequences from trajectories and pairs parameterized programs with natural-language guidance for execution and adaptation~\cite{wang2026webxskill}. Complementing these synthesis efforts, The Tool Illusion systematically examines whether tools yield consistent gains and how tool design and side effects shape web-agent performance across tool sources, models, frameworks, and benchmarks~\cite{lou2026toolillusion}. Together, these works show how tools can be created and why tool-set quality matters, but leave open how to construct and maintain an effective inventory from a large candidate pool; moreover, some API discovery methods do not explicitly ensure or evaluate coverage. In contrast, \proj{} begins with comprehensive, systematically collected trajectories to establish broad behavioral coverage, then filters, deduplicates, and reselects the derived tools to maintain a compact, high-utility inventory as user demand changes.

\subsection{Meta-agent optimization}

Selecting a capability inventory offline is insufficient when the requests presented to an agent change over time. Meta-agent optimization offers a broader view of adaptation by treating artifacts surrounding a foundation model as objects that can be improved. GEPA evolves prompts through reflection on execution trajectories and retains complementary candidates along a Pareto frontier~\cite{agrawal2026gepa}. AlphaEvolve applies a similar iterative pattern to executable programs, using domain-specific evaluators to select promising modifications~\cite{novikov2025alphaevolve}. Meta-Harness broadens the optimized artifact to model-harness code by giving an outer-loop proposer access to source code, evaluation scores, and prior execution traces~\cite{lee2026metaharness}. Together, these works show that an agentic system can improve without updating the base model's weights, but they optimize a prompt, program, or harness rather than the membership of a deployed tool set. In contrast, \proj{} adapts the capability inventory itself: it uses observed requests, tool usage, fallback invocations, and failures to decide which tools to add, retain, or remove. This changes what the agent can invoke directly as user demand evolves, rather than only changing how a fixed set of capabilities is used.

\paragraph{Research space.}
Trajectory generation supplies candidate behaviors; tool and skill synthesis renders them executable; and meta-agent optimization enables adaptation. However, prior work does not jointly address capability inventory optimization: selecting which synthesized capabilities to deploy and adapting that set as demand changes. \proj{} addresses this gap by constructing a quality-controlled, deduplicated, and utility-ranked MCP tool portfolio offline and reselecting deployed tools online based on observed coverage gaps and usage, thereby treating compactness, functional coverage, deployment cost, and user alignment as first-class objectives.

\section{System Architecture}
\label{sec:system_architecture}

\textsc{\proj{}} uses a multi-agent architecture that converts recorded web trajectories into a set of compact and efficient APIs and adjusts the deployed API set based on runtime demand. See Figure \ref{fig:system_diagram} for the system workflow. Within this pipeline, the \emph{Offline Selection module} generates and filters trajectory-derived APIs, while the \emph{Online Dynamic Reselection module} identifies coverage gaps and updates the active API set. Together, these modules transform an automatically generated, potentially redundant, and non-user-aligned API pool into a compact, demand-responsive capability inventory, reducing the agent's active tool space while restoring coverage as new needs emerge.

\subsection{Offline Selection Module}
\label{subsec:offline_selection}

The offline module transforms collected trajectories into executable browser-automation APIs and organizes them through two selection filters inspired by prior work about high quality tool-set composition~\cite{lou2026toolillusion}. The initial API set, $S_{\mathrm{init}}$, contains all generated APIs. The Quality Filter (QF) removes APIs with unsuitable granularity and functional duplicates, producing $S_{\mathrm{QF}}$, while the Usage Likelihood Filter (ULF) retains high-utility APIs while preserving capability diversity, producing $S_{\mathrm{ULF}}$. The $S_{\mathrm{ULF}}$ set initializes the online system, whereas $S_{\mathrm{QF}}$ serves as the candidate pool for later adaptation.

\begin{figure}
    \centering
    \includegraphics[width=0.5\linewidth]{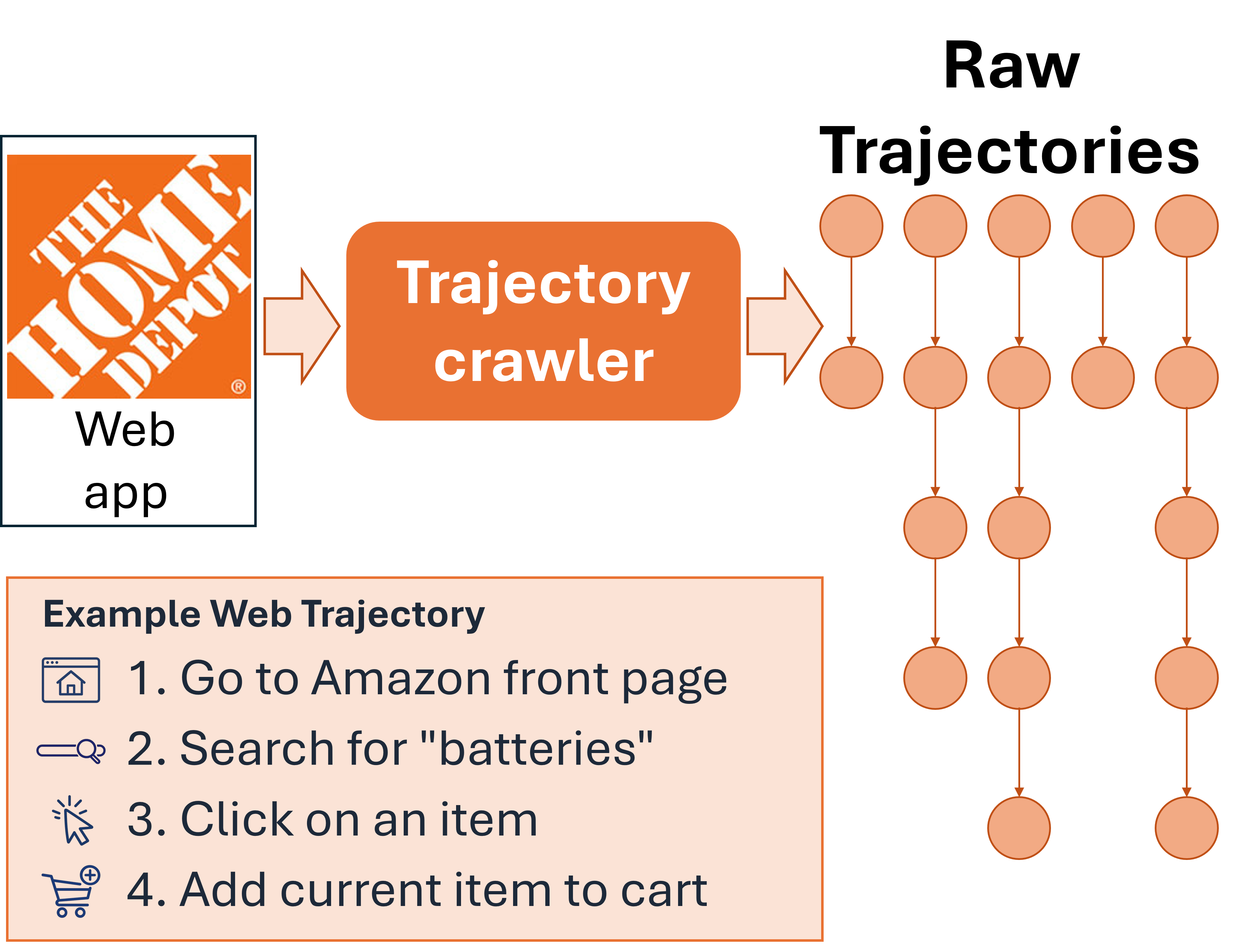}
    \caption{\textbf{Overview of trajectory crawling.} Given a target website, the trajectory crawler systematically explores available interactions and collects diverse multi-step web trajectories, each represented as an ordered sequence of actions. The resulting corpus captures trajectories of varying lengths and task structures; the inset illustrates an example trajectory consisting of navigation, search, item selection, and cart interaction.}
    \label{fig:traj_crawling}
\end{figure}

\paragraph{Trajectory Crawling.}
The offline pipeline starts with a comprehensive set of web agent trajectories obtained through crawling methods proposed in prior work \cite{faisal2026autosurfer}. Each trajectory contains a task description, contextual information, and an ordered sequence of browser actions (Figure~\ref{fig:traj_crawling}).

\paragraph{Capability (API) Construction.}
Then, a code generation LLM is prompted to convert each trajectory into an asynchronous function that executes UI actions through Playwright API \cite{microsoft_playwright}. The generator replaces scenario-specific values, such as user names, search terms, titles, and free-form text, with semantic function parameters. Fixed interface labels, built-in options, and URLs remain literal. Consequently, \emph{each generated program represents a reusable task template rather than a single recorded execution}. The generator also ensures that a function's name, signature and docstring specify the task, arguments, execution steps, and return value.

\paragraph{Quality Filter (QF).}
As the first stage of API selection, we apply a filter with two components: granularity filtering and functional deduplication. We first use a metadata-derived granularity score to assess each trajectory's abstraction level (Appendix~\ref{appendix:task-granularity-evaluation-prompt}), removing trajectories that are overly specific, low-level, or excessively fine-grained, and retain reusable, user-facing operations at the target abstraction level. We then remove functionally redundant APIs to reduce the agent's action space and tool-selection ambiguity. Specifically, we embed each API's parameterized docstring and cluster semantically related APIs using greedy centroid-based clustering with a cosine-similarity threshold of 0.75. Within each cluster, pairs with similarity above 0.80 are evaluated by an LLM verifier that compares task descriptions, action sequences, and executable code for functional equivalence (Appendix~\ref{appendix:functional-deduplication-detection-prompt}). For each verified duplicate pair, we retain the more complete trajectory, breaking ties by fewer interaction steps and then earlier corpus order. The resulting collection forms $S_{\mathrm{QF}}$, with cluster assignments preserved for subsequent coverage-aware selection.

\paragraph{Usage Likelihood Filter (ULF).}
As the second stage of API selection, ULF prioritizes capabilities that are likely to be useful in practice while preserving semantic diversity. An LLM assigns each API in $S_{\mathrm{QF}}$ a usage-likelihood score from 1 to 5, conditioned on the target website and the parameterized API description (Appendix~\ref{appendix:usage-likelihood-evaluation-prompt}). APIs scoring 4 or 5 are retained by default. To prevent the removal of entire capability categories, ULF additionally retains the highest-scoring API from any otherwise uncovered semantic cluster, provided its score is at least 3. This criterion prioritizes high-demand APIs while preserving moderately useful capabilities when needed for coverage. The resulting collection forms $S_{\mathrm{ULF}}$.

\paragraph{MCP Capability Set Synthesis.}
The selected programs are exposed as executable capabilities through a dynamic MCP server. At initialization, the server scans the configured capability bank and registers each program as an MCP tool. The program's function name is used as the tool identifier, its docstring defines the tool description, and its semantic parameters are translated into the corresponding input schema. At runtime, the MCP server serves as the execution interface between a web agent (the client) and the selected capability set. The agent receives user tasks and invokes a tool by specifying its task-relevant arguments, and the server resolves the corresponding program, supplies the required execution context, and executes the capability. This design converts the selected program bank into a structured and directly callable MCP tool set, providing the agent with a uniform interface to reusable web capabilities.

Importantly, this capability set is not treated as static. While the offline selection process provides an initial bank of high-quality and broadly useful tools, the user needs encountered during deployment may differ from those represented in the initial corpus. The \proj{} system therefore also provides a natural interface for modifying the available capability set at runtime. Building on this interface, the following section introduces an online dynamic adaptation module that updates the tool set in response to capabilities required during agent interaction.

\subsubsection{Online Dynamic Reselection (DR)}
\label{subsec:online_dynamic_reselection}

The active API set is initialized from $S_{\mathrm{ULF}}$, while $S_{\mathrm{QF}}$ serves as a broader pool of quality-filtered and deduplicated candidates that can be introduced as new capability needs emerge. Adaptation proceeds through four steps: runtime usage logging, coverage-gap detection, candidate retrieval and capability addition, and capability pruning.

\paragraph{Logged Runtime Usage.}
After each user task during online usage, the system records the user intent, tool-call sequence, task outcome, and cumulative API usage. These logs provide both task-level evidence for identifying missing capabilities and usage statistics for maintaining the active API set.

\paragraph{Coverage-Gap Detection.}
A task is marked as uncovered when the active capabilities are insufficient to complete it. This occurs either when the synthesized APIs fail a task or when the runtime web agent determines that no suitable API is available for the current user task. The system maintains uncovered tasks within a rolling window and normalizes their intents by parameterizing task-specific values. The normalized intents are then embedded and clustered by semantic similarity to identify recurring capability gaps. Reselection is triggered periodically or after a configurable number of uncovered tasks have accumulated. These parameters are deployment-dependent; in our evaluation, we use a rolling window of 100 tasks and a cosine-similarity threshold of 0.82 for clustering.

\paragraph{Candidate Retrieval and Capability Addition.}
When a \texttt{reselection} is triggered, for each unresolved coverage-gap cluster, an LLM is given the representative normalized intents together with descriptions of inactive APIs from $S_{\mathrm{QF}}$. It selects up to five candidate APIs whose functionality can address the observed gap (Appendix~\ref{appendix:dynamic-api-reselection-prompt}). Clusters for which no suitable candidate is identified remain unresolved and are reconsidered after additional runtime evidence is collected. Selected candidates are added to the active capability set and become available for future tasks as the MCP server reloads its registered tools.

\paragraph{Capability Pruning.}
Addition and pruning are governed independently: new capabilities are introduced in response to observed coverage gaps, whereas removal requires sustained evidence of non-use. APIs that have been invoked are protected from pruning, while never-used APIs become eligible for removal only after a configurable inactivity period. Newly added APIs are similarly protected by a grace period to allow sufficient opportunity for use, and pruning is constrained by a minimum active-set size. These parameters are also deployment-dependent; in our evaluation, unused APIs become eligible for pruning after 50 tasks, newly added APIs receive a 50-task grace period, and at least 10 APIs are retained in the active set.

This asymmetric adaptation policy favors rapid expansion in response to observed coverage gaps while requiring sustained evidence of non-use before removing existing capabilities. Consequently, the deployed API set tracks the runtime task distribution while remaining compact and preserving capabilities that demonstrate practical utility.

\section{Evaluation}
\label{sec:eval}

\subsection{Resulting API sets}

\proj{} progressively constructs the API set exposed to the worker agent. 
The initial API set $S_{\mathrm{init}}$ contains all 1,283 trajectory-derived APIs. 
QF produces $S_{\mathrm{QF}}$, containing 215 APIs after granularity filtering and deduplication, and ULF produces $S_{\mathrm{ULF}}$, containing 87 APIs after usage-likelihood filtering. 
Finally, DR is calibrated (\ref{sec:DR-calibration}) to adapt the deployed capability set according to observed task demand, API usage, and coverage gaps, producing $S_{\mathrm{DR}}$, containing 33 APIs in our evaluation. 
These configurations correspond to the successive API-selection stages evaluated below.

\subsection{Dynamic Calibration Protocol}

\label{sec:DR-calibration}

We perform the dynamic reselection calibration by randomizing all 106 WebArena Postmill tasks and treating the resulting order as a simulated user interaction sequence. 
As tasks are processed, the system adapts the deployed API set based on observed task demand, API usage, and detected coverage gaps. 
This calibration produces the final $S_{\mathrm{DR}}$ set, which is frozen and used in the following evaluation. 
Our evaluation focuses on the quality of the resulting calibrated capability set rather than intermediate adaptation behavior; consequently, we do not report per-step calibration accuracy, trigger counts, task-family statistics, or intermediate API-set measurements.
This protocol assumes that the user's task interests and their relative frequencies remain approximately stable after calibration. 
Future work will further evaluate the system with real world dynamic usage data.

\subsection{Evaluation on WebArena-Postmill}

\label{sec:functional-eval}

\paragraph{\proj{} agent configuration.}
The evaluation setup comprises two agents, both instantiated with Copilot using Claude Opus 4.8~\cite{anthropic2026claudeopus48}. The \emph{tooling agent} is connected to our system's MCP server, which exposes the selected capability API set and, when enabled by the experimental condition, a function for invoking the fallback agent. The \emph{fallback agent} is connected to Playwright MCP~\cite{microsoft_playwright_mcp}. Given a user task, the tooling agent reasons about which APIs from the current capability set to invoke, supply the required arguments, and sequence multiple API calls when needed to accomplish the task. If the available APIs are insufficient and ReAct fallback is enabled, the tooling agent invokes the fallback agent that iteratively reasons over the webpage state and executes low-level browser actions until completion or termination. This configuration uses compact, reusable APIs for supported tasks while retaining general browser-interaction coverage through ReAct. If the condition is "without ReAct", the tooling agent does not have access to this fallback agent.

\paragraph{Benchmark.}
We evaluate \proj{} on all 106 tasks from the WebArena \cite{zhou2024webarenarealisticwebenvironment} Postmill benchmark using four trajectory-derived MCP API sets: $S_{\mathrm{init}}$, $S_{\mathrm{QF}}$, $S_{\mathrm{ULF}}$, and the dynamically calibrated $S_{\mathrm{DR}}$. Each set is evaluated with and without the ReAct fallback: the ReAct-enabled setting measures end-to-end agent performance, whereas the no-ReAct setting isolates the functional coverage and execution efficiency of the synthesized APIs. We additionally report \texttt{ReAct only (reference)}, in which the agent completes tasks through ReAct without a trajectory-derived API set.

\paragraph{Metrics.}
Correctness is the average benchmark score across all 106 tasks. 
Latency, MCP API calls, LLM calls, and component-level token counts are reported as all-task/successful-task averages. 
We separately report measurements for the tooling agent and the fallback agent. 
For ReAct-enabled configurations, \texttt{Total Request Token Cost} and \texttt{Total Completion Token Cost} report the aggregate token consumption across both components.

\begin{table*}[t]
  \centering
  \small
  \setlength{\tabcolsep}{2.5pt}

  \caption{
  Functional evaluation with the ReAct Fallback on all 106 WebArena Postmill tasks, comparing the static API-selection stages with the dynamically reselected capability set.
  \texttt{ReAct only} provides a ReAct-only reference without a trajectory-derived API set.
  ``All'' and ``Succ.'' denote averages over all tasks and successful tasks, respectively.
  Bold indicates the best value across all evaluated conditions for each metric and averaging scope -
  higher is better for correctness and the number of MCP API calls; lower is better for all other metrics.
  }
  \label{tab:reselected-with-react}

  \resizebox{\textwidth}{!}{%
  \begin{tabular}{
      l
      >{\columncolor{gray!4}}r
      >{\columncolor{gray!10}}r
      !{\hspace{5pt}}
      >{\columncolor{gray!4}}r
      >{\columncolor{gray!10}}r
      !{\hspace{5pt}}
      >{\columncolor{gray!4}}r
      >{\columncolor{gray!10}}r
      !{\hspace{5pt}}
      >{\columncolor{gray!4}}r
      >{\columncolor{gray!10}}r
      !{\hspace{5pt}}
      >{\columncolor{gray!4}}r
      >{\columncolor{gray!10}}r
  }

    \toprule

    &
    \multicolumn{2}{c}{\shortstack{\texttt{ReAct only}\\(reference)}}
    &
    \multicolumn{2}{c}{\shortstack{$S_{\mathrm{init}}$\\w/ ReAct}}
    &
    \multicolumn{2}{c}{\shortstack{$S_{\mathrm{QF}}$\\w/ ReAct}}
    &
    \multicolumn{2}{c}{\shortstack{$S_{\mathrm{ULF}}$\\w/ ReAct}}
    &
    \multicolumn{2}{c}{\shortstack{$S_{\mathrm{DR}}$\\w/ ReAct}}
    \\

    \cmidrule(lr){2-3}
    \cmidrule(lr){4-5}
    \cmidrule(lr){6-7}
    \cmidrule(lr){8-9}
    \cmidrule(lr){10-11}

    Metric
    & \textbf{All} & \textbf{Succ.}
    & \textbf{All} & \textbf{Succ.}
    & \textbf{All} & \textbf{Succ.}
    & \textbf{All} & \textbf{Succ.}
    & \textbf{All} & \textbf{Succ.}
    \\
    \midrule

    Correctness
    & 0.875 & 1
    & 0.880 & 1
    & 0.835 & 1
    & 0.873 & 1
    & \textbf{0.906} & 1
    \\

    Avg Task Latency (s)
    & 102.63 & 96.65
    & 93.31 & 72.90
    & 85.97 & \textbf{64.29}
    & 86.91 & 75.51
    & \textbf{72.45} & 69.17
    \\

    Avg \# MCP API Calls
    & 1.09 & 1.08
    & 1.34 & 1.23
    & 1.73 & 1.63
    & \textbf{1.76} & 1.67
    & 1.67 & \textbf{1.68}
    \\

    Avg \# Tooling LLM Calls
    & \textbf{3.04} & \textbf{3.07}
    & 3.52 & 3.39
    & 4.09 & 3.95
    & 4.41 & 4.12
    & 3.98 & 3.90
    \\

    Avg Tooling Request Tokens
    & \textbf{29.70k} & \textbf{29.94k}
    & 2.39m & 2.30m
    & 403.30k & 387.98k
    & 205.13k & 188.89k
    & 98.46k & 95.89k
    \\

    Avg Tooling Completion Tokens
    & \textbf{1.00k} & \textbf{0.99k}
    & 1.93k & 1.76k
    & 1.56k & 1.41k
    & 1.72k & 1.53k
    & 1.43k & 1.40k
    \\

    Avg \# Fallback LLM Calls
    & 11.64 & 11.92
    & \textbf{4.21} & 4.17
    & 4.25 & \textbf{3.67}
    & 5.33 & 5.04
    & 4.82 & 4.69
    \\

    Avg Fallback Request Tokens
    & 3.10m & 3.18m
    & \textbf{970.96k} & 962.51k
    & 1.01m & \textbf{875.03k}
    & 1.31m & 1.24m
    & 1.22m & 1.18m
    \\

    Avg Fallback Completion Tokens
    & 2.21k & 2.20k
    & \textbf{0.79k} & 0.79k
    & 0.84k & \textbf{0.69k}
    & 1.00k & 0.97k
    & 0.98k & 0.92k
    \\

    Avg Total Request Token Cost
    & 3.13m & 3.21m
    & 3.36m & 3.26m
    & 1.42m & \textbf{1.26m}
    & 1.52m & 1.43m
    & \textbf{1.32m} & 1.28m
    \\

    Avg Total Completion Token Cost
    & 3.21k & 3.19k
    & 2.72k & 2.55k
    & \textbf{2.40k} & \textbf{2.10k}
    & 2.72k & 2.50k
    & 2.41k & 2.32k
    \\

    \bottomrule

  \end{tabular}%
  }
\end{table*}

\begin{table*}[t]
  \centering
  \small
  \setlength{\tabcolsep}{2.5pt}

  \caption{
  Functional evaluation without ReAct fallback on all 106 WebArena Postmill tasks. Annotations same as Table\ref{tab:reselected-with-react}.}
  \label{tab:reselected-without-react}

  \resizebox{\textwidth}{!}{%
  \begin{tabular}{
      l
      >{\columncolor{gray!4}}r
      >{\columncolor{gray!10}}r
      !{\hspace{5pt}}
      >{\columncolor{gray!4}}r
      >{\columncolor{gray!10}}r
      !{\hspace{5pt}}
      >{\columncolor{gray!4}}r
      >{\columncolor{gray!10}}r
      !{\hspace{5pt}}
      >{\columncolor{gray!4}}r
      >{\columncolor{gray!10}}r
  }

    \toprule

    &
    \multicolumn{2}{c}{\shortstack{$S_{\mathrm{init}}$\\w/o ReAct}}
    &
    \multicolumn{2}{c}{\shortstack{$S_{\mathrm{QF}}$\\w/o ReAct}}
    &
    \multicolumn{2}{c}{\shortstack{$S_{\mathrm{ULF}}$\\w/o ReAct}}
    &
    \multicolumn{2}{c}{\shortstack{$S_{\mathrm{DR}}$\\w/o ReAct}}
    \\

    \cmidrule(lr){2-3}
    \cmidrule(lr){4-5}
    \cmidrule(lr){6-7}
    \cmidrule(lr){8-9}

    Metric
    & \textbf{All} & \textbf{Succ.}
    & \textbf{All} & \textbf{Succ.}
    & \textbf{All} & \textbf{Succ.}
    & \textbf{All} & \textbf{Succ.}
    \\
    \midrule

    Correctness
    & \textbf{0.604} & 1
    & 0.599 & 1
    & 0.502 & 1
    & 0.601 & 1
    \\

    Avg Task Latency (s)
    & 99.11 & 58.03
    & 84.48 & 41.98
    & 86.85 & 44.30
    & \textbf{79.50} & \textbf{36.93}
    \\

    Avg \# MCP API Calls
    & 1.93 & \textbf{1.44}
    & 1.76 & 1.42
    & 1.71 & 1.43
    & \textbf{1.99} & \textbf{1.44}
    \\

    Avg \# Tooling LLM Calls
    & \textbf{4.05} & 3.58
    & 4.27 & 3.85
    & 4.35 & 4.10
    & 4.66 & \textbf{3.51}
    \\

    Avg Tooling Request Tokens
    & 2.78m & 2.43m
    & 452.58k & 383.82k
    & 225.07k & 191.24k
    & \textbf{142.86k} & \textbf{90.11k}
    \\

    Avg Tooling Completion Tokens
    & 3.57k & 2.04k
    & 3.44k & 1.48k
    & 3.16k & 1.57k
    & \textbf{2.92k} & \textbf{1.40k}
    \\

    \bottomrule

  \end{tabular}%
  }
\end{table*}

\subsection{Results}

\paragraph{Results with ReAct Fallback.}
Table~\ref{tab:reselected-with-react} shows that the static filtering stages primarily trade tool-context size against dependence on the ReAct Fallback.
Among the static configurations, $S_{\mathrm{init}}$ achieves the highest correctness (88.0\%), but incurs substantial tooling overhead, requiring 2.39m request tokens per task on average.
The first filtering stage reduces this overhead by nearly an order of magnitude, to 403.30k tooling request tokens, while also lowering average task latency from 93.31\,s to 85.97\,s.
This efficiency gain is accompanied by a decrease in correctness from 88.0\% to 83.5\%, suggesting that granularity filtering and deduplication remove substantial redundancy but also eliminate capabilities useful for a subset of tasks.

The usability-based second filter further reduces the tooling context to 205.13k request tokens while recovering correctness to 87.3\%.
However, this smaller static capability set relies more heavily on ReAct execution: average ReAct LLM calls increase from 4.25 under $S_{\mathrm{QF}}$ to 5.33 under $S_{\mathrm{ULF}}$.
Consequently, the reduction in tooling cost does not translate directly into lower end-to-end cost; average total request-token cost increases slightly from 1.42m to 1.52m.
This result highlights a limitation of aggressive static filtering: reducing the exposed tool set can simplify tool selection while shifting unresolved tasks to the ReAct Fallback.

DR yields a more favorable tradeoff.
The calibrated $S_{\mathrm{DR}}$ set achieves the highest correctness (90.6\%) and lowest all-task latency (72.45\,s), while reducing tooling request tokens to 98.46k and average total request-token cost to 1.32m.
Compared with $S_{\mathrm{init}}$, this represents more than a $24\times$ reduction in tooling request tokens while improving correctness.
These results suggest that runtime usage signals allow the system to retain capabilities that are important for the observed task distribution without reintroducing the large tool context of the full capability bank.
Thus, dynamic adaptation improves both effectiveness and efficiency relative to purely static API selection.

\paragraph{Results without ReAct.}
Table~\ref{tab:reselected-without-react} isolates the functional coverage provided directly by the synthesized APIs.
QF nearly preserves the correctness of $S_{\mathrm{init}}$, decreasing only from 60.4\% to 59.9\%, while reducing average task latency from 99.11\,s to 84.48\,s and tooling request tokens from 2.78m to 452.58k.
This result indicates that granularity filtering and deduplication remove substantial redundancy from the original capability bank while preserving most of its direct task coverage.

The second filter produces a still smaller tool context, reducing tooling request tokens to 225.07k, but correctness falls to 50.2\%.
The 9.7-percentage-point decrease relative to $S_{\mathrm{QF}}$ shows that ULF alone does not fully preserve functional coverage: APIs estimated to be less frequently useful may nevertheless be essential for less common task types.
The contrast with its stronger ReAct-enabled performance further indicates that the ReAct Fallback compensates for capabilities removed during static utility filtering.

DR largely recovers this lost coverage.
The frozen $S_{\mathrm{DR}}$ set achieves 60.1\% correctness, nearly matching both $S_{\mathrm{init}}$ (60.4\%) and $S_{\mathrm{QF}}$ (59.9\%), while requiring only 142.86k tooling request tokens and reducing average task latency to 79.50\,s.
Thus, adaptation does not simply expand the active set toward $S_{\mathrm{init}}$; instead, it identifies a substantially smaller capability set that better matches the observed runtime distribution.
Taken together with the ReAct-enabled results, these findings support the intended role of online reselection: static filtering provides a compact initial capability set, while runtime adaptation selectively restores capabilities whose practical value becomes evident during deployment.

\section{Discussion, Limitations, and Future Work}
\label{sec:discussion-future-work}

\subsection{Web Agent Capability Management As a First-Class Component}
Our results indicate that tool-set composition has a substantial effect on both the efficiency and effectiveness of web agents. Static filtering can remove large amounts of redundant tooling context, while dynamic reselection recovers capabilities that become important under observed usage. Our results suggest that efficient, user-aligned management of web-agent tools may be an important complement to improvements in planning and reasoning. We view \proj{} as an early exploration of this design space through a meta-agent capability-management framework. Future work could move beyond per-user adaptation by learning shared capability profiles across groups of users with similar task distributions, while still allowing individual specialization where needed.

\subsection{Evaluation Realism Through Real World Usage}
Our current evaluation uses templated WebArena tasks to simulate repeated user interaction. This provides a controlled setting for measuring coverage, latency, and tooling cost, but does not fully capture real-world usage. Human requests are often noisier, less regular, and more context-dependent than benchmark templates, while benchmark suites may also contain tasks that occur rarely in practice. Consequently, the observed benefit of dynamic reselection should be interpreted as evidence that the framework can adapt to a stable simulated task distribution, rather than as a complete characterization of production behavior. A key direction for future work is therefore to evaluate \proj{} with real user interactions over longer time horizons, including changes in user interests, heterogeneous task frequencies, and naturally occurring failures. Such studies would enable a more realistic assessment of whether capability reselection improves end-to-end efficiency and task success under deployment conditions.

\subsection{Verification for Compact Capability Sets}
Reliability remains a major challenge for web agents, particularly when individual actions can have irreversible or safety-sensitive effects. Stronger reliability mechanisms, including formal verification or exhaustive validation of action sequences, can be prohibitively expensive when applied online to open-ended browser interaction. The compact, programmatic capability sets produced by \proj{} offer a complementary direction: because frequently used behaviors are distilled into a relatively small set of reusable APIs, these capabilities can potentially be inspected, tested, or formally verified offline before deployment. This shifts part of the reliability burden from repeated online reasoning to pre-validated executable actions. Future work could therefore combine dynamic capability management with offline verification, policy checking, or sandboxed testing, enabling the system to retain the flexibility of web agents while increasing the reliability and safety of commonly executed operations.

\bibliographystyle{johd}
\bibliography{bib}


\appendix

\section{Technical appendices and supplementary material}
\label{appendix}

\subsection{Task Granularity Evaluation Prompt}
\label{appendix:task-granularity-evaluation-prompt}

\begin{promptbox}
{\small\sffamily
You are an expert in evaluating web agent task trajectories. Your task is to assess the task granularity and coherence of the function that a trajectory represents.

A trajectory consists of the following components:
\begin{enumerate}
  \item \textbf{High-level Instruction:} Describes the user's intended task (e.g., ``Search for `AskReddit' and review the results'').
  \item \textbf{Action History:} A sequence of actions performed by the agent, including the action type (e.g., click, fill), followed by the description of the particular action.
\end{enumerate}

We define the ``task granularity'' of a function (web trajectory) as how well it corresponds to a single, meaningful, self-contained user task.
\begin{itemize}
  \item If granularity is too low: the function is too trivial or atomic (e.g., only navigating to a page), and does not accomplish a meaningful goal.
  \item If granularity is too high: the function combines multiple loosely related steps, includes unnecessary transitions, or contains actions not required for the core task.
\end{itemize}

A well-formed function should:
\begin{itemize}
  \item Achieve a clear user goal.
  \item Avoid unnecessary intermediate steps.
  \item Not be decomposable into smaller meaningful sub-tasks without losing usefulness.
\end{itemize}

When evaluating specificity, consider these key aspects:

\textbf{Evaluate based on:}
\begin{enumerate}
  \item \textbf{Goal completeness}---does the trajectory achieve a meaningful outcome?
  \item \textbf{Minimality}---are all steps necessary for that outcome?
  \item \textbf{Focus}---does it avoid unrelated or loosely related sub-tasks?
\end{enumerate}

\textbf{Special rule for user-ask trajectories:}
\begin{itemize}
  \item If the last action in Action History is a \texttt{generate\_answer} action, treat the requested answer as successfully retrieved in that final step.
  \item In this case, do not mark the trajectory as low granularity or incomplete solely because there are no specific steps to generate this answer.
\end{itemize}

\textbf{Scoring Guidelines:}

Rate the function (web trajectory) on a scale of 1--5 based on task granularity and coherence.

\textbf{5 = Severely over-scoped (too coarse)}
\begin{itemize}
  \item Combines multiple distinct or loosely related tasks.
  \item Contains unnecessary transitions or exploratory steps.
  \item Includes actions not required for a single clear goal.
  \item Can be cleanly split into multiple meaningful sub-functions.
\end{itemize}

\textbf{4 = Slightly over-scoped}
\begin{itemize}
  \item Has a clear main goal, but includes extra or redundant steps.
  \item Some actions are not strictly necessary for achieving the goal.
  \item Could be simplified without losing functionality.
\end{itemize}

\textbf{3 = Well-balanced (ideal)}
\begin{itemize}
  \item Represents a single, coherent, meaningful user task.
  \item Includes all and only the necessary steps.
  \item Has no redundant navigation or unrelated sub-tasks.
  \item Cannot be meaningfully decomposed further without losing usefulness.
\end{itemize}

\textbf{2 = Slightly under-scoped}
\begin{itemize}
  \item Partially achieves a goal but stops short of a meaningful outcome.
  \item Is missing key steps required to complete the task.
  \item Feels like a fragment of a larger task.
\end{itemize}

\textbf{1 = Severely under-scoped (too atomic)}
\begin{itemize}
  \item Only performs trivial or mechanical actions (e.g., navigation, clicking).
  \item Does not accomplish a meaningful user goal.
  \item Is clearly just a step within a larger task.
\end{itemize}

You are given the following information:
\begin{enumerate}
  \item High-level Instruction: \texttt{\{\{HIGH\_LEVEL\_INSTRUCTION\}\}}
  \item Action History with action for each step: \texttt{\{\{ACTION\_HISTORY\}\}}
\end{enumerate}

Your response should comply with the following JSON schema:

\begin{quote}
\footnotesize\ttfamily
\{\\
\hspace*{1em}"reasoning": <brief explanation of the granularity score>,\\
\hspace*{1em}"score": <integer from 1 to 5>\\
\}
\end{quote}
}
\end{promptbox}

\subsection{Functional Deduplication Detection Prompt}
\label{appendix:functional-deduplication-detection-prompt}

\begin{promptbox}
{\small\sffamily
You are an expert in evaluating web-automation trajectories. Your task is to determine whether two trajectories are functional duplicates.

Two trajectories are functional duplicates if they accomplish the same user task in essentially the same way. When comparing them, ignore:
\begin{itemize}
  \item Superficial differences in wording.
  \item Differences in the ordering of equivalent steps.
  \item Differences in specific input values.
\end{itemize}

You are given the following information:

\textbf{Trajectory A:}
\begin{quote}
\small\ttfamily
id: \{\{TRAJECTORY\_A\_ID\}\}\\
description: \{\{TRAJECTORY\_A\_DESCRIPTION\}\}\\
actions:\\
\{\{TRAJECTORY\_A\_ACTIONS\}\}\\
code:\\
\{\{TRAJECTORY\_A\_CODE\}\}
\end{quote}

\textbf{Trajectory B:}
\begin{quote}
\small\ttfamily
id: \{\{TRAJECTORY\_B\_ID\}\}\\
description: \{\{TRAJECTORY\_B\_DESCRIPTION\}\}\\
actions:\\
\{\{TRAJECTORY\_B\_ACTIONS\}\}\\
code:\\
\{\{TRAJECTORY\_B\_CODE\}\}
\end{quote}

Determine whether Trajectory A and Trajectory B are functional duplicates and explain your judgment briefly. Your response should comply with the following JSON schema:

\begin{quote}
\footnotesize\ttfamily
\{\\
\hspace*{1em}"is\_duplicate": <true or false>,\\
\hspace*{1em}"reasoning": <brief explanation of the judgment>\\
\}
\end{quote}
}
\end{promptbox}

\subsection{Usage Likelihood Evaluation Prompt}
\label{appendix:usage-likelihood-evaluation-prompt}

\begin{promptbox}
{\small\sffamily
You are an expert in evaluating web-automation trajectories. Your task is to rate the ``usage likelihood'' of a trajectory: how likely a real user would want to perform the task on the given website.

\textbf{Scoring Guidelines:}

Rate the trajectory on a scale of 1--5 and select the single best-fitting integer.

\textbf{5 = Very likely / core popular goal}
\begin{itemize}
  \item A primary reason people visit the website; extremely common.
  \item Examples for Reddit include ``Search for a topic,'' ``Post a comment on a submission,'' and ``Upvote a post.''
\end{itemize}

\textbf{4 = Likely / common goal}
\begin{itemize}
  \item A frequent, natural task, though not the single most central one.
  \item Examples include ``Sort a forum's submissions by New,'' ``Subscribe to a forum,'' and ``Edit your own comment.''
\end{itemize}

\textbf{3 = Plausible / occasional goal}
\begin{itemize}
  \item Something a fair number of users do sometimes, but not routinely.
  \item Examples include ``Update your profile biography,'' ``Create a new post in a niche forum,'' and ``Save a post for later.''
\end{itemize}

\textbf{2 = Unlikely / niche goal}
\begin{itemize}
  \item A task performed only by power users or under specific circumstances; rarely done.
  \item Examples include ``Change notification email frequency,'' ``Toggle a rarely used display preference,'' and ``Block a specific user.''
\end{itemize}

\textbf{1 = Very unlikely / unnatural goal}
\begin{itemize}
  \item A rare settings adjustment or an artificial or contrived intention that almost no real user would form.
  \item Examples include ``Flip an obscure advanced setting no one changes'' and ``A task combining unrelated steps a real user would never chain.''
\end{itemize}

You are given the following information:
\begin{enumerate}
  \item Website: \texttt{\{\{WEBSITE\_DESCRIPTION\}\}}
  \item Parameterized trajectory description: \texttt{\{\{TRAJECTORY\_DESCRIPTION\}\}}
\end{enumerate}

Rate this trajectory's usage likelihood and explain your judgment briefly. Your response should comply with the following JSON schema:

\begin{quote}
\footnotesize\ttfamily
\{\\
\hspace*{1em}"rating": <integer from 1 to 5>,\\
\hspace*{1em}"reasoning": <brief explanation of the rating>\\
\}
\end{quote}
}
\end{promptbox}

\subsection{Dynamic API Reselection Prompt}
\label{appendix:dynamic-api-reselection-prompt}

\begin{promptbox}
{\small\sffamily
You are an expert in selecting web-automation APIs. Your task is to select candidate APIs that can resolve a group of user tasks that the current API tool set could not solve because no available API matched them.

The unresolved tasks are grouped by template. Each \texttt{template} line represents a task pattern whose parameters are masked using the following placeholders:
\begin{itemize}
  \item \texttt{<val>} represents a value such as a name, description, or quoted text.
  \item \texttt{<list>} represents a list of values.
  \item \texttt{<num>} represents a number.
\end{itemize}
Each template is followed by a few concrete \texttt{e.g.} examples. Use the templates to reason about structural intent while using the examples as grounding. Parameter values supplied to an API may differ from those in the examples.

You are also given a catalog of candidate APIs. Each catalog entry contains an API identifier and its docstring description.

When selecting APIs:
\begin{itemize}
  \item Choose only APIs that, when invoked with appropriate parameter values, genuinely accomplish one or more of the unresolved task templates.
  \item Select at most \texttt{\{\{MAX\_APIS\_PER\_RESELECTION\}\}} APIs.
  \item Return selected APIs in preference order.
  \item Use each API identifier verbatim as it appears in the candidate catalog.
  \item If none of the candidate APIs apply, return an empty list.
\end{itemize}

You are given the following information:

\textbf{User task templates the current API set could not solve:}
\begin{quote}
\small\ttfamily
\{\{UNRESOLVED\_TASK\_TEMPLATES\}\}
\end{quote}

\textbf{Candidate APIs (id: description):}
\begin{quote}
\small\ttfamily
\{\{CANDIDATE\_API\_CATALOG\}\}
\end{quote}

Determine which API identifiers can solve these tasks. Your response should comply with the following JSON schema:

\begin{quote}
\footnotesize\ttfamily
\{\\
\hspace*{1em}"solving\_apis": [<API identifiers in preference order>],\\
\hspace*{1em}"reasoning": <brief explanation of the selection>\\
\}
\end{quote}
}
\end{promptbox}

\end{document}